\documentclass{article}

\usepackage{PRIMEarxiv}

\usepackage[utf8]{inputenc} 
\usepackage[T1]{fontenc}    
\usepackage{hyperref}       
\usepackage{url}            
\usepackage{booktabs}       
\usepackage{amsfonts}       
\usepackage{nicefrac}       
\usepackage{microtype}      
\usepackage{lipsum}
\usepackage{fancyhdr}       
\usepackage{graphicx}       
\graphicspath{{media/}}     
\usepackage{amsbsy}

\usepackage{amssymb}
\usepackage{enumitem}
\usepackage{csquotes}
\usepackage[dvipsnames]{xcolor}
\usepackage{rotating}
\usepackage{colortbl}
\usepackage{tcolorbox}
\usepackage{tikz}
\usepackage{hyperref}  
\hypersetup{hidelinks}
\usepackage{subfigure}

\title{A Brief Survey on Representation Learning based Graph Dimensionality Reduction Techniques}

\author{
  Akhil Pandey Akella \\
  Dept. of Computer Science \\
  Northern Illinois University \\
  Dekalb, IL\\
  \texttt{\{aakella\}@niu.edu}
}

\begin{document}
\maketitle

\begin{abstract}
Dimensionality reduction techniques map data represented on higher dimensions onto lower dimensions with varying degrees of information loss. Graph dimensionality reduction techniques adopt the same principle of providing latent representations of the graph structure with minor adaptations to the output representations along with the input data. There exist several cutting edge techniques that are efficient at generating embeddings from graph data and projecting them onto low dimensional latent spaces. Due to variations in the operational philosophy, the benefits of a particular graph dimensionality reduction technique might not prove advantageous to every scenario or rather every dataset. As a result, some techniques are efficient at representing the relationship between nodes at lower dimensions, while others are good at encapsulating the entire graph structure on low dimensional space. We present this survey to outline the benefits as well as problems associated with the existing graph dimensionality reduction techniques. We also attempted to connect the dots regarding the potential improvements to some of the techniques. This survey could be helpful for upcoming researchers interested in exploring the usage of graph representation learning to effectively produce low-dimensional graph embeddings with varying degrees of granularity.
\end{abstract}


\section{Introduction}

\begin{figure*}[!h]
 \centering
 \includegraphics[width=0.80\textwidth]{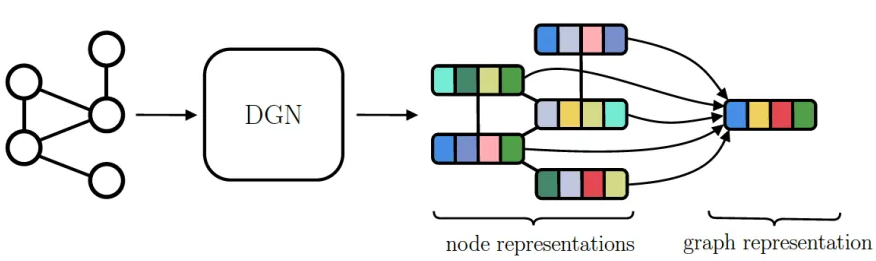}
 \caption{
The figure shows an architecture overview \cite{Yuan2020} of the dimensionality reduction technique applied to graph data to produce embedding in a different latent space. The Deep Graph Network (DGN) takes a graph structure as an input and can capture the information at the node as well as the graph. The resulting output can produce node level and graph level representations.
 }
 \label{fig:overview}
\end{figure*}

\begin{figure*}[t]
\centering
\begin{tabular}[t]{l} 
\subfigure[Input Graph $G_1$]{\label{fig:ig} 
	\includegraphics[width=0.16\linewidth]{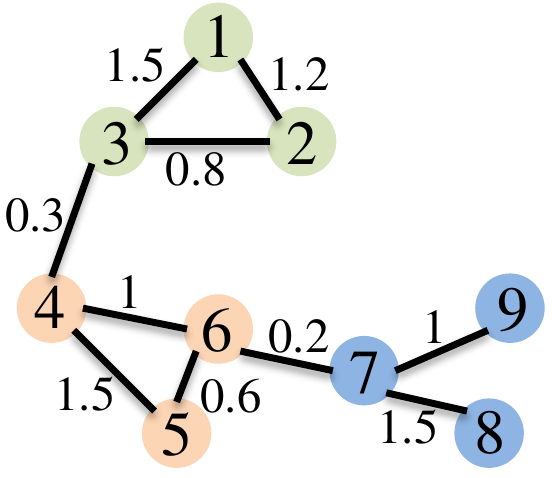}}
\subfigure[Node Embedding]{\label{fig:ne}
	\includegraphics[width=0.195\linewidth, height=3.15cm]{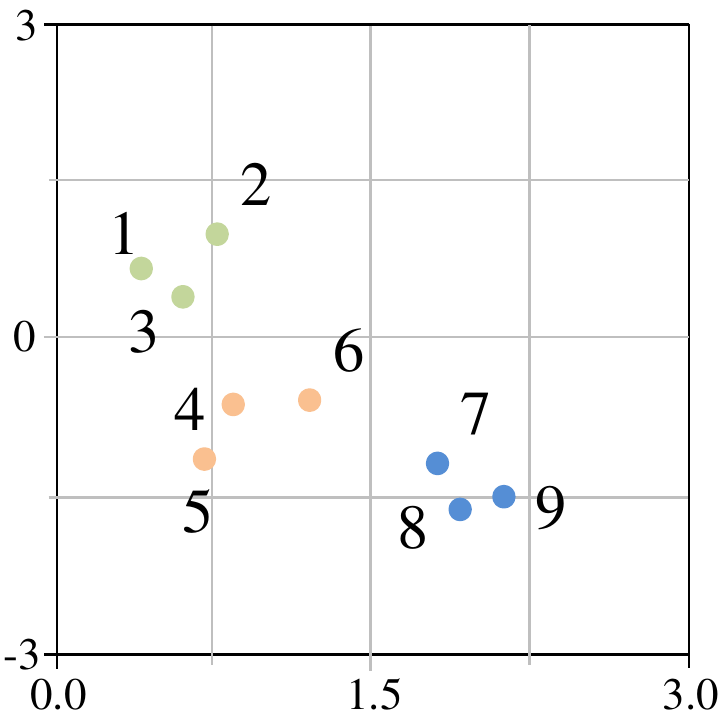}}
\subfigure[Edge Embedding]{\label{fig:ee}
	\includegraphics[width=0.195\linewidth, height=3.15cm]{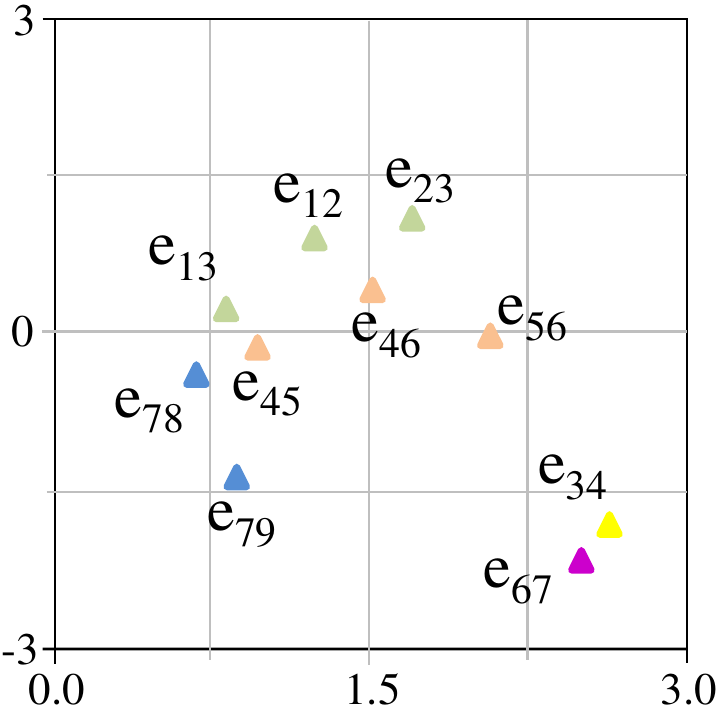}}
\subfigure[Substructure Embedding]{\label{fig:se}
	\includegraphics[width=0.2\linewidth, height=3.15cm]{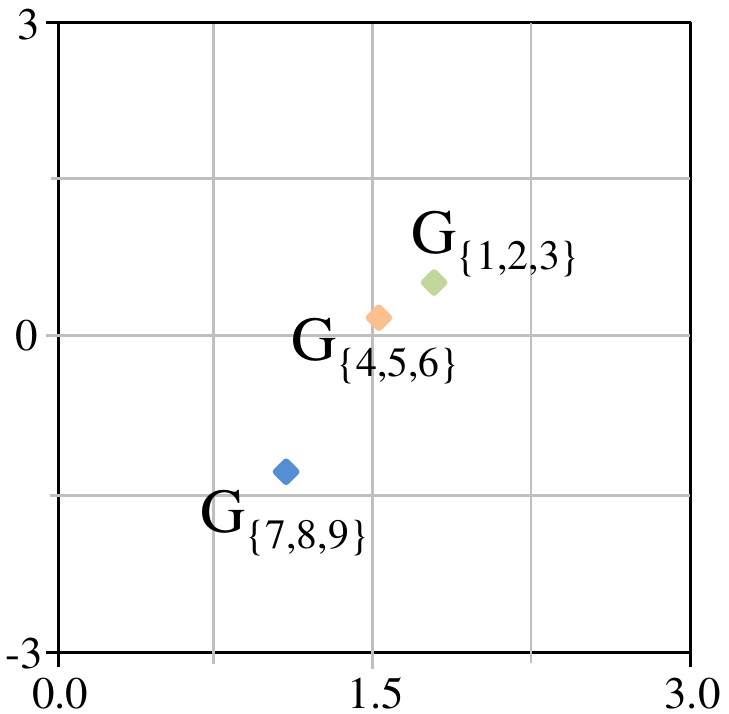}}
\subfigure[Whole-Graph Embedding]{\label{fig:we}
	\includegraphics[width=0.2\linewidth, height=3.15cm]{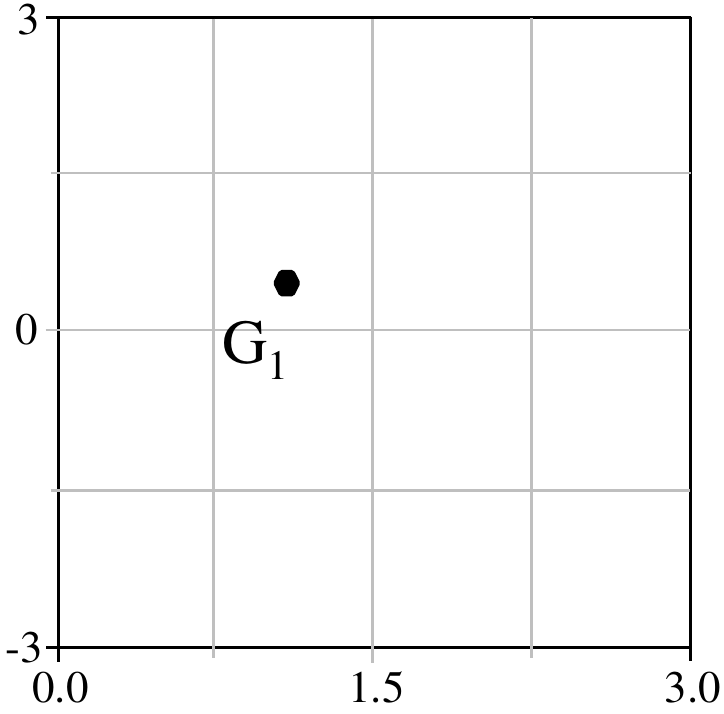}}
\end{tabular}
\vspace{-3mm}
\caption{Example of a graph representation being mapped onto latent spaces with varying degrees of granularity and information by \cite{Cai2018}.}
\label{fig:graph-embed-example}
\end{figure*}

Graph representation learning as a concept has grown over the past years giving room for the deep learning community to make significant advances in diverse problems relating to networks or graphs such as graph embedding, node classification, graph clustering, and many more. Graphs as data representations used in multiple situations spanning different disciplines in science, and some of the prominent examples include citation networks \cite{Zhou2018, Wu2020}, social networks \cite{Zhou2018, Wu2020}, dataflow graphs, e-commerce user interest graphs, knowledge graphs \cite{Zhou2018, Wu2020}, drugs  \cite{Zhou2018}. The aforementioned problems related to graphs are somewhat prevalent in all of the examples mentioned above, and the application of deep learning techniques on the said variations of the graphs has given room for the concept of graph representation learning to grow. The overarching trend in the last decade was the confluence of deep learning as a problem-solving strategy onto different problem spaces. It has played a pivotal role in offering solutions with significantly high levels of efficacy. Observing the fundamental principle behind deep learning, i.e. gradient-descent based learning \cite{LeCun2015},  one can notice the adaptability of the concept on graph data with minor architectural modifications. We must also notice how the underlying non-euclidean nature of the graph representations allows it to enclose complex relationships between the entities that are not necessarily possible on euclidean spaces. The graphs are also capable of encoding additional information at the node level in the form of features. This degree of intricacy allows the graphs to encompass composite information that is not usually possible on the euclidean space.

The idea behind representation learning \cite{Hamilton2018} approaches is simple: employing Machine learning(ML)/Deep Learning(DL) models to learn graph representations such that the model can map nodes, or entire (sub)graphs, as points in a low-dimensional latent space,  i.e., dimensionality reduction. Traditionally, the legacy dimension reduction techniques \cite{Sorzano2014} were based out of euclidean spaces and the fundamental principle used for achieving it was matrix decompositions \cite{Goodfellow2016, Sorzano2014}. Representation learning can be useful here in transforming the legacy approach behind graph dimensionality reduction, thereby allowing the possibility to map different components of graphs (nodes, features, edges) onto low dimensional latent spaces. The resulting output of the operation would give us embeddings commonly referred to as \textbf{Graph Embeddings} \cite{Cai2018}. Fig. \ref{fig:overview} highlights this same process using a Deep Graph Network (DGN) \cite{Yuan2020} or a Graph Neural Network \cite{Wu2020} to produce node embeddings or graph embeddings. 

Furthermore, as highlighted in \cite{Cai2018}, it would be a misnomer to identify the task of producing these graph embeddings as somewhat trivial. The primary reason being aware of the type of the graph we are dealing with, and the relationships it encompasses. The normative definition of a graph would give us the idea that all graphs are nearly the same in terms of the structure they hold, when it’s quite not the case. The different variations of the graphs \cite{Cai2018} include homogeneous graphs, heterogeneous graphs, graphs from non-relational data. While these could be grouped as different variants of input, there are also different variants of output. Fig. \ref{fig:graph-embed-example} shows the different output representations possible for a given input graph $G$, namely \textit{node embedding \cite{Grover2016, Abu-El-Haija2018, Škrlj2019, Epasto2019}, edge embedding, subgraph/substructure embedding, and complete graph embedding \cite{Goyal2018-a, Pan2019}}. Due to these key differences it is difficult to hypothesize mathematical models that have the capacity to contextually understand what is the information that ought to be preserved from higher dimensions. Similar to legacy dimensionality reduction techniques, there are also legacy graph representation learning techniques as highlighted by \cite{Hamilton2018} that include latent space models, manifold learning algorithms, geometric deep learning techniques \cite{Bronstein2017}. Although there is vast literature encompassing the diverse legacy graph representation learning techniques, the scope of the paper is limited to advancements made in graph representation learning using ML/DL techniques \cite{Cao2016}.

The past couple of years saw an uptick in deep learning techniques applied on this problem space \cite{Cao2016} to produce efficient graph embeddings and the progress is partially facilitated using model architectures like Graph Autoencoders, Graph Convolutional Networks, Generative Graph Neural Networks \cite{Wu2020}. Each of these architectures have a niche spectrum of representation learning tasks that they perform efficiently which include :

\begin{itemize}
	
	\item \textbf{Node level} --- node classification, semi-supervised learning, clustering tasks, recommendation tasks, detecting communities.
	\item \textbf{Edge level} --- reconstruction of graphs, predicting the missing links in a graph..
	\item \textbf{Graph level} --- discovering patterns in graphs, information visualization, graph classification.
	
\end{itemize}

In order for the models to perform these tasks at node level, or edge level, or at graph level having the graph embeddings is a crucial prerequisite to begin with since most of these tasks are just downstream machine learning problems \cite{Hamilton2018}, and Graph Visualization tasks. As mentioned above, the challenges associated with producing these embeddings on low dimensional latent spaces exist on multiple fronts. If we move past the issues related to the problem setting \cite{Cai2018}, the next set of problems linger around resource constraints and time consume for the models to understand the neighbourhood features \cite{Hamilton2018}. Lastly, as mentioned in \cite{Bengio2014} the main noticeable challenge of producing embeddings from graph representations using  representation learning is the one associated with the cost function of the model or rather the objective of the training. This challenge still persists and is regarded as an open question \cite{Bengio2014} although it doesn’t limit the capacity of the models in any way to extract underlying relationships from higher dimensions and map them onto lower dimensions. 

\section{Background}
\label{sec:background}

We came up with three facets that would help us evaluate the current models in the literature. The facets mentioned below helped us frame certain questions that were answered in the Table. \ref{table:papers}. It is to be noted that in Table \ref{table:papers} we included only the most prominent works since alternatives with minor variations exist on a larger scale.

\subsection{Facets associated with evaluating related work}
\begin{itemize}
	\item[\textbf{1.}]
	\textbf{Model:} A model that outputs graph representations is efficient when it is capable of producing embeddings on different datasets. Additionally, efficiency as a metric should not come at a cost of not being able to infer the model. Consequently, observing if the model consumes significant time to generate the output representation and placing emphasis on time taken to learn the representations on higher dimensions are crucial for ensuring the overall fitness of the model.
	
	\item[\textbf{2.}]
	\textbf{Scalability:} Scalability of dimensionality reduction techniques is always a difficult thing to ensure because of the way the graph representations are transformed. We were keen to observe and note to what degree are the models producing graph representations scalable. Are they scalable if there are new vertices that are ought to be assigned to a local neighborhood, Are they scalable in a graph that changes continuously, and lastly, Are they scalable when graphs dynamically update information at the nodes ?
	
	\item[\textbf{3.}]
	\textbf{Improvement:} Once we build the models to generate embeddings there is always scope for improvement. The improvement could happen on different fronts. We were interested to observe if an existing work could be improved either on the: Computational front, visual front, or on the side resource management. Our motivation to include these sub-criteria is based on our thought that the models which produce better graph embeddings must be capable of combining the technical efficiency (computational efficiency, and resource consumption) with meaninful visualizations.

\end{itemize}

\subsection{Our contributions and related work}
There are three important paradigms to our survey that made our analysis different from the existing literature. First, our survey attempts to connect the vast literature on representation learning \cite{Cai2018, Hamilton2018} for graph structured data in order to produce graph embeddings. The methods dimensionality reduction techniques \cite{Sorzano2014}, and architectural improvements to existing techniques. Additionally, our survey is different from the ones mentioned above as we join the dots starting from introducing a technique, highlighting the pros/cons, and providing the literature for potential improvements on different fronts such as time-complexity, scalability, and resource consumption.

\begin{table}[!bht]
	\caption{Commonly used notations for a Graph \cite{Wu2020}.}
	\label{tab:graph-notations}
	\centering
	\begin{tabular} {  p{3cm} p{7cm} } \toprule \hline
		\textbf{Notations}& \textbf{Descriptions} \\ \midrule \hline
		$G$& A graph. \\ \hline
		$V$& The set of nodes in a graph.\\ \hline
		$v$ & A node $v\in V$. \\ \hline
		$E$& The set of edges in a graph.\\ \hline
		$e_{ij}$ & An edge $e_{ij}\in E$.\\ \hline
		$\mathbf{A}$ & The graph adjacency matrix.  \\ \hline
		$n$ & The number of nodes, $n = |V|$. \\ \hline
		$m$ & The number of edges, $m = |E|$. \\ \hline
		$d$  & The dimension of a node feature vector.\\ \hline
		$b$ & The dimension of a hidden node feature vector. \\ \hline
		$c$  & The dimension of an edge feature vector.\\ \hline
		$\mathbf{X} \in \mathbf{R}^{n\times d}$ & The feature matrix of a graph. \\ \hline
		$\mathbf{x} \in \mathbf{R}^n$ & The feature vector of a graph in the case of $d=1$. \\ \hline
		$\mathbf{x}_v \in \mathbf{R}^d$ & The feature vector of the node $v$. \\ \hline
		$\mathbf{X}^e \in \mathbf{R}^{m\times c}$ & The edge feature matrix of a graph. \\ \hline
		$\mathbf{x}^e_{(v,u)} \in \mathbf{R}^{c}$ & The edge feature vector of the edge $(v,u)$. \\ \hline
		$\mathbf{X}^{(t)} \in \mathbf{R}^{n\times d}$ & The node feature matrix of a graph at the time step $t$. \\
		\bottomrule
	\end{tabular}
\end{table}

\section{Methods}
\label{sec:methodology}

As articulated in the introduction, dimensionality reduction techniques on graphs can include \textit{node based embeddings, edge based embeddings feature based embeddings, whole-graph based embeddings}. Our interest for this survey primarily lies in unravelling the literature behind techniques for producing feature based embeddings,  graph based embeddings, and lastly hybrid approaches that attempt to encapsulate both feature level representations and whole graph representation. The literature highlighted in the survey would include the crucial components of the published work, the potential for improvements, and lastly the effectiveness of the visualization component associated with the output representations.

\subsection{Notations, and definitions}
Before we begin summarizing the novel works on the said categories, there are certain presuppositions and notations that we would like to put forward so that there is a contextual agreement on the terminology used in the paper. In the context of graphs, we define a graph $G$ to be a representation that consists of a set of vertices or nodes $V$, and edges $E$. Here, $v_i \in V$ represents the $i^{th}$ node. If we have a connection between two nodes such that there is an edge pointing from $v_j$ and $v_i$ then the edge $e_{ij} \in E$ is noted as $e_{ij} = (v_i, v_j)$. The total number of nodes $V$ in a given graph $G$ are represented using $n$. Similarly, the total edges in graph $G$ are given by $m$. $A$ represents the adjacency matrix of size $nxn$ that is useful to us in presenting information about nodes that are adjacent to each other. The values in the matrix $A$ are 1 if $e_{ij} \in E$ and 0 if $e_{ij} \notin E$. Additional notations that include the Feature matrix of the entire graph $G$, feature matrix of a node $v$ are clearly denoted in Table. \ref{tab:graph-notations}. Lastly, unless specified when we refer to a graph $G$, usually we are referring to graphs that are not weighted.
Along with the mentioned notations there are certain definitions that could help readers new to this topic to have a contextual understanding of the term. \textit{First-order proximity} is defined as the pairwise distance between two vertices of a graph. Similarly, \textit{Second-order proximity} is defined as the pairwise distance between a vertex and its neighborhood structure.

\begin{figure*}[!h]
 \centering
 \includegraphics[width=0.80\textwidth]{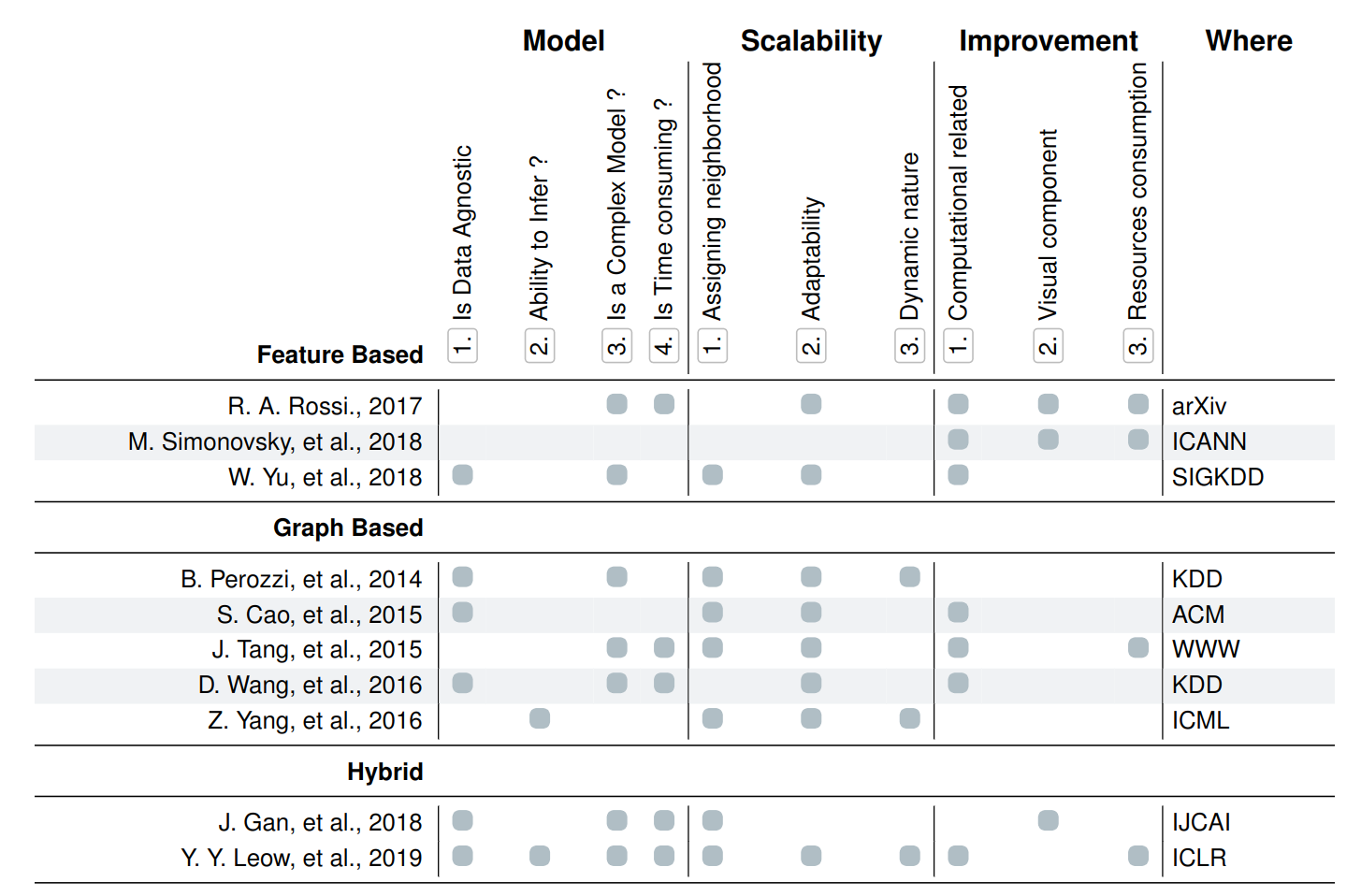}
 \caption{
Overview of the categories to which the literature used for the survey belong. Works that provide dimensionality reduction for feature representation of a graph come under Feature Based methods. Similarly, literature covering representation learning techniques purely for graphs fall under Graph based methods. Lastly, hybrid approaches include published works that attempt to produce a combination of output representations.
 }
 \label{fig:papers}
\end{figure*}

\subsection{Feature Based}

We regard a technique to be \textbf{feature based} when the primary motivation of the technique is to produce feature representations of the vertices for a given input graph. As we know from Table. \ref{tab:graph-notations} the feature matrix $X$ is a matrix with $n X d$ dimensionality. This would mean that the dimensionality reduction applied on the feature representations are euclidean in nature. Although the source of the information has changed, i.e feature matrix $X$ from vertices $V$, there is no reason to believe that the features obtained from graph data would be treated any differently. This survey on dimensionality reduction techniques \cite{Sorzano2014} outlines all of the different linear and non-linear variants of the existing techniques. For this reason, we would like to shed light on some of the new and promising techniques like DeepGL \cite{Rossi2017}, and different variants of ML/DL learning techniques \cite{Min2010, Gutiérrez-Gómez2019} like auto-encoders \cite{Simonovsky2018, Yu2018, Kipf2016}.
\begin{figure*}[!h]
	\centering
	\includegraphics[width=0.70\textwidth]{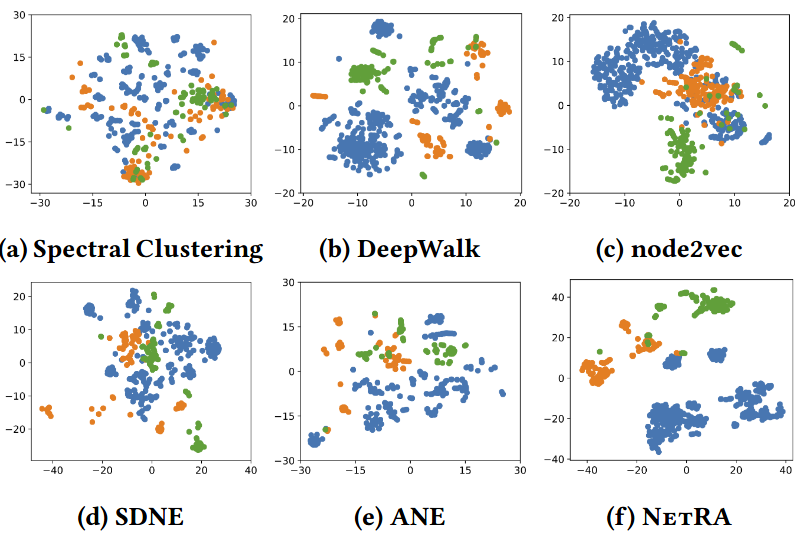}
	\caption{
		Qualitative assesment of feature based representations \cite{Yu2018} that employ DL techniques like autoencoders to produce embeddings for the JDK network repository dataset.
	}
	\label{fig:featurebased-comparison-1}
\end{figure*}

One of the prominent DL technique that produces low dimensional feature representations with minimal loss is auto-encoder. Auto-encoders are an unsupervised deep neural networks that are capable of learning how to encode the information onto lower dimensions such that, when reconstructed using decoders there is minimal or no loss of information. The learning process of the auto-encoders \cite{Yu2018} can be identified with the following cost function : 
\begin{equation}
\min E_{x ~ P_{data}(x) [dist(x, h_{\psi}(f_{\phi}(x))) ]}
\label{eq:auto-encoder}
\end{equation}

In Eq. \ref{eq:auto-encoder} $h_{\psi}$ refers to the decoder of the auto-encoder while $f_{\phi}$ refers to the encoder of the auto-encoder, $x \in R^n$ is the input given to the auto-encoder. The issues concerning auto-encoders is related to complexity of the models and the infrastructure required to ensure the auto-encoder models are built. The models comprise millions of parameters and when compared with other non-linear dimensionality reduction techniques they appear to be complex and huge. Additionally, most of the auto-encoders require good hardware to ensure that the models are built and trained in least amount of time possible. Although these are hurdles if addressed then we can observe how the visualization component as seen in Fig. \ref{fig:featurebased-comparison-1} of these output representations is promising when compared with other prominent methods.

\subsection{Graph Based}

We regard a technique to be \textbf{graph based} when the primary motivation of the technique is to produce graph embeddings that represent all of connections in the network. Additionally, the crucial aspect of these techniques is make sure that the model is capable of generating output embedding that efficiently maps graph data onto low dimensional latent spaces.

The prominent examples of graph based methods that don't necessarily use representation learning techniques usually begin with Laplacian Eigenmaps(LE) \cite{Cai2018, Hamilton2018}. LE are a classical example of non-linear manifold learning techniques that use an objective function to minimize, $\sum_{ij} = ||y_i - y_j||^2$ with $W_{ij} = tr(Y^{T}LY)$; the term $y$ represents the output embeddin vector, and $W_{ij}$ represents weight matrix that stores information about essential elements. There is one straightforward advantage to this approach, due to the motivation partially placed on geometry this manifold learning approach \cite{Sorzano2014} preserves connections in the local neighborhood. Although it preserves the nodes in the local neighborhood there is the disadvantage of having disconnected graphs being formed as a result of using LE. Additionally, the visual representation of the embeddings results in close proximity clusters that wouldn't be helpful for us to understand the different classes in the dataset.

Moving away from the legacy approaches, there is dearth of literature that focuses on providing graph based embeddings using representation learning techniques. The promising work here includes Structural Deep Network Embedding(SDNE) \cite{Wang2016}, DeepWalk \cite{Perozzi2014}, LINE \cite{Tang2015}, GraRep \cite{Cao2015}, and tsNET \cite{Kruiger2017}.

\begin{figure*}[!h]
	\centering
	\includegraphics[width=0.70\textwidth]{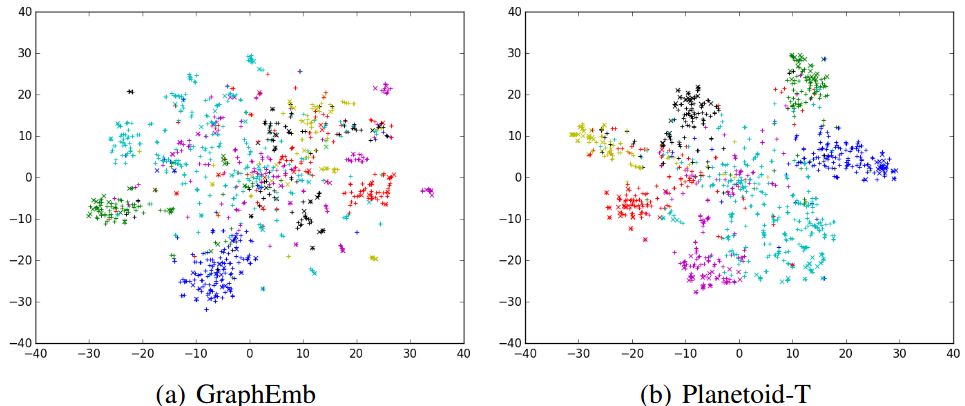}
	\caption{
		Qualitative assesment of graph based models \cite{Yang2016} that employ semi-supervised learning techniques to produce graph embeddings on the CORA dataset. The GraphEmb here refers to nothing but a graph emebeddings generated using \cite{Perozzi2014}.
	}
	\label{fig:graphbased-comparison-1}
\end{figure*}

\begin{figure*}[!h]
	\centering
	\includegraphics[width=0.99\textwidth]{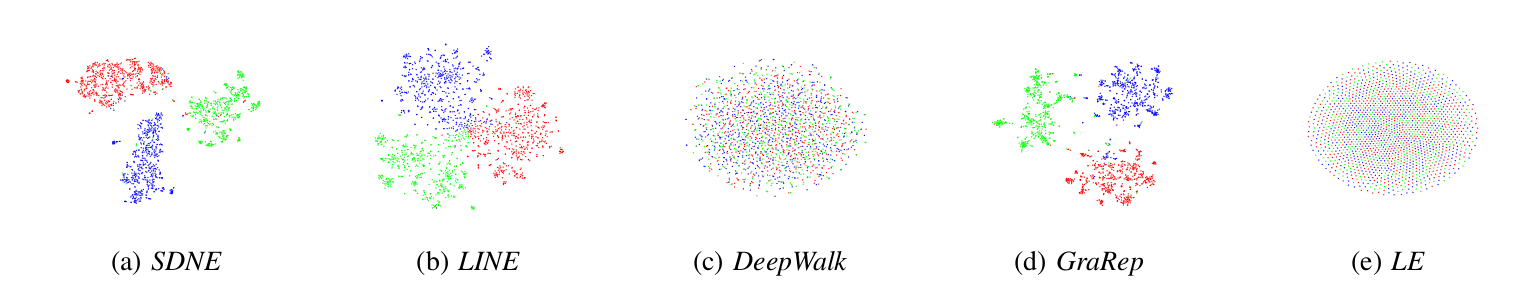}
	\caption{
		Qualitative assesment of graph based models \cite{Wang2016} that employ ML/DL techniques to produce output representations of graphs. All of these visualizations were performed on the 20 Newsgroup dataset. 
	}
	\label{fig:graphbased-comparison-2}
\end{figure*}

First, let us observe DeepWalk, which is nothing but a DL model trying to use representation learning to produce graph embeddings for a given graph. But there is more to it than just DL being applied, for starters it is based on a problem called \textins{Random Walk}, which is choosing a random vertex $v_i$ from the root vertex. As highlighted in \cite{Perozzi2014}, fusion of random walks with representation learning with provide us with an opportunity to produce network embeddings that are both adaptive, and continuous. Adaptability with regards to understanding the neighborhood even in the case of an evolving graph. Continuity in the sense of providing a clearly spaced latent representations of vertices on a continuous space. The model builds binary tree $T$ from the given vertices $V$ of a graph $G$ and then later applies random walks to select different vertices $v$ and lastly applies SkipGram model to generate output representations $\phi \in R^{|V| X d}$. Therefore, the primary disadvantages of the model is that it can easily produce graph embeddings for networks having binary edges. Fig. \ref{fig:graphbased-comparison-1}a.) shows the visualization of the output embedding on CORA dataset. Similarly, Fig. \ref{fig:graphbased-comparison-2} c.) shows output embedding on the 20 Newsgroup dataset.

Moving on, let us understand SDNE, which is a good example of a technique that uses semi-supervised learning \cite{Goodfellow2016} to build graph embeddings and ensure at the same time that the local structure of the input graph is preserved. The structure preservation aspect is dictated using a hyperparameter that ensures a balance between first-order proximity and second-order proximity. SDNE uses semi-supervised learning to obtain a initial set of parameters and then uses the obtained parameters to optimize the output representations. Adopting this strategy will help the model while back propagating the activations, as it has an idea as to what are the optimum parameters to learn. This leaves us with the problem of adaptation to new vertices, i.e if we have a new vertex that has no connections to the existing vertices of the graph then there is no way to ensure a neighborhood is assigned to the new vertex.

Usually the DL/ML models employ different variants of Gradient Descent to reduce the time taken to compute the representation and Stochastic-Gradient Descent is the most chosen function. But LINE chooses a different optimization function citing effectiveness and efficacy as the reason \cite{Tang2015}. Moving further, there are some advantages of using LINE, primarily it could scale to millions of vertices on the graphs thereby making sure that there is no issue of assigning neighborhoods to new vertices. Also, it can provide inference on the relationships observed by the model using an edge-sampling strategy. Lastly, the technicalities and the advantages mentioned would mean nothing if the output representations are not meaningful. Fig. \ref{fig:graphbased-comparison-2} b.) shows the visualization of output embeddings produced for 20 Newsgroup dataset using LINE.

Another prominent example of a method that use semi-supervised learning to produce graph embeddings would be \textit{Planetoid} \cite{Yang2016}. Due to the choice of semi-supervised learning, the core advantage of Planetoid is its ability to generate embeddings from graph data either inductively or transductively. The transductive mode operates via. making predictions on unlabeled samples without generalizing to unobserved instances. The inductive mode on the other hand is akin to classifying unobserved instances. Additionally, planetoid doesn't stress on the usage of graph laplacian for regularization \cite{Goodfellow2016} since they regularize the model learning the labels without actually regularizing the learning that takes place on graph embeddings. Lastly, Fig. \ref{fig:graphbased-comparison-1} b.) shows the transductive mode of planetoid applied on the CORA dataset. 

\subsection{Hybrid}

We regard a technique to be \textbf{hybrid} when it captures both the feature representations at the node combined with the graph on lower dimensions. To our knowledge there are only two prominent examples Robust Graph Dimensionality Reduction (RGDR) \cite{Gan2018}, and GraphTSNE \cite{Leow2019} fits in this category as the remaining contemporary techniques do not attempt to fuse different approaches. Let us understand the intricacies behind these techniques along with why they are considered to be hybrid in the first place.
\begin{figure*}[!h]
	\centering
	\includegraphics[width=0.80\textwidth]{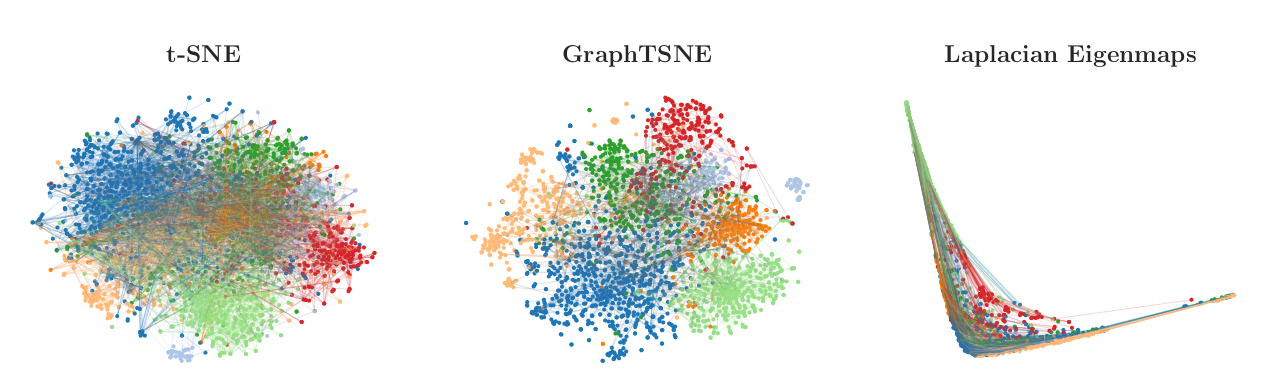}
	\caption{
		The figure shows comparison of graphTSNE with t-SNE and Laplacian Eigenmaps when applied on the CORA dataset. 
	}
	\label{fig:graphtsne-comparison}
\end{figure*}

Robust Graph Dimensionality Reduction(RGDR) \cite{Gan2018} learns both the transformation matrix and the graph matrix on the intrinsic space which is constructed by the refined original data. Here transformation matrix is nothing but a transformation of the input feature matrix $X$ on the intrinsic space. The function described in Eq. \ref{eq:rgdr} was used as an objective function for transforming the initial matrices. Once the transformation matrices are obtain, they are used for training in some of the classification tasks mentioned in \cite{Gan2018}. Additional to the primary task of performing good on the classification tasks, experimental analysis of RDGR on binary and multi-class classification datasets proved that RDGR is capable of converging quickly to learn the representations of the transformed graph matrix and the feature matrix. The intricate mathematical nature of the study didn't give room for a visual component to RGDR since the motivation for the approach had basis in what we referred in the introduction section as downstream ML tasks. Due to this reason, the effectiveness and fitness of the technique was quantitative in nature.

\begin{equation}
\min_{U, U^TU = I} \sum_{i, j = 1}^{n} a_{ij} || U_{x_i} - U_{x_j}||^{2}_{2}
\label{eq:rgdr}
\end{equation}

Similarly, the reason GraphTSNE model is regarded as hybrid is because it uses a parametric model like Graph Convolutional Network \cite{Kipf2017, Wu2020} to leverage representation learning techniques for producing output graph representations. In other words, the technique tries to adopt that include both the given graph $G$ and the node features $V$. 

\begin{equation}
h^{l+1}_{i} = ReLU(U^{l}h^{l}_{i} + \frac{1}{n(i)} \sum_{j \to i} \eta_{ij} \odot V^{l}h^{l}_{j}) + h^{l}_{i}
\label{eq:graphtsne}
\end{equation}

The core working philosophy of GraphTSNE is quite similar to t-SNE \cite{Maaten2008} but the changes in the cost function and the propagation rule allow it to gain complete advantage of graph inputs. Eq. \ref{eq:graphtsne} shows the layer wise propagation rule that is used for the model. Here, $h^{l}_{i}$ represents the latent representation of the node $v_i$ at the layer $l$, i.e this propagation rule dictates how information in the hidden layers is being updated with respect to the given input. Additionally, since the model was performing both graph representation along with feature representation there was a modified sub-loss function that was used to calculate the distance matrix. Specifically, the distance matrix $D$ had information about how far are the point-pairs. The experimentation for the paper \cite{Leow2019} was based purely on graph data sets like CORA \footnote{https://relational.fit.cvut.cz/dataset/CORA}, CITESEER \footnote{https://linqs-data.soe.ucsc.edu/public/citeseer-mrdm05/} \cite{Rossi2015}, and Pubmed\footnote{https://linqs-data.soe.ucsc.edu/public/Pubmed-Diabetes.tgz} \cite{Rossi2015}.

The main improvements that could be offered to this model lean towards the computational aspect of the model. GraphTSNE \cite{Leow2019} takes $O(N^2)$ time to compute the distance matrix $D$ and it could be reduced to $O(NlogN)$ using tree based algorithms like \cite{Maaten2014}.

Lastly, visualizations of this model clearly show their effectiveness in terms of proximity each neighborhood holds along with the ability to reflect the higher level relationships effectively for the whole graph. Fig. \ref{fig:graphtsne-comparison} serves as a qualitative assessment of the GraphTSNE model in terms of how it compares with legacy dimensionality reduction techniques \cite{Sorzano2014}, along with graph only representations.

\section{Conclusion}
\label{sec:conclusion}

We presented a brief survey that outlines the prominent literature covering dimensionality reduction techniques for graph-based data using representation learning techniques based out of ML/DL. Our survey outlined the new literature of feature-based representations, graph-based representations, and hybrid approaches as well. Along with highlighting the prominent literature we have also judged them based on three facets: \textbf{Model, Scalability, and Improvement}. Finally, the future direction of this survey would be aligned towards expansion of the literature relating to graph-based methods and also potentially new hybrid approaches.

\bibliographystyle{unsrt}

\begin{thebibliography}{1}

\bibitem{Zhou2018}
J. Zhou et al.,
``Graph Neural Networks: A Review of Methods and Applications,''
\emph{arXiv:1812.08434 [cs, stat]}, Jul. 2019, Accessed: Apr. 05, 2020. [Online]. Available: http://arxiv.org/abs/1812.08434.

\bibitem{LeCun2015}
Y. LeCun, Y. Bengio, and G. Hinton,
``Deep learning,''
\emph{Nature}, vol. 521, pp. 436--444, 2015.

\bibitem{Goodfellow2016}
I. Goodfellow, Y. Bengio, and A. Courville,
``Deep Learning,''
\emph{MIT Press}, 2016, [Online]. Available: http://www.deeplearningbook.org.

\bibitem{Bronstein2017}
M. M. Bronstein, J. Bruna, Y. LeCun, A. Szlam, and P. Vandergheynst,
``Geometric deep learning: going beyond Euclidean data,''
\emph{IEEE Signal Process. Mag.}, vol. 34, no. 4, pp. 18–42, 2017.

\bibitem{Kipf2017}
T. N. Kipf, and M. Welling,
``Semi-Supervised Classification with Graph Convolutional Networks,''
\emph{International Conference on Learning Representations (ICLR) - ICLR `2017}, 2017.

\bibitem{Rossi2015}
R. A. Rossi and N. K. Ahmed.
"The Network Data Repository with Interactive Graph Analytics and Visualization", 
{\em AAAI}, 2015.


\bibitem{Cai2018}
H. Cai, V. W. Zheng, and K. C.-C. Chang,
``A Comprehensive Survey of Graph Embedding: Problems, Techniques and Applications,''
\emph{arXiv:1709.07604 [cs]}, Feb. 2018, Accessed: Feb. 29, 2020. [Online]. Available: http://arxiv.org/abs/1709.07604.

\bibitem{Wu2020}
Z. Wu, S. Pan, F. Chen, G. Long, C. Zhang, and P. S. Yu,
``A Comprehensive Survey on Graph Neural Networks,''
\emph{IEEE Trans. Neural Netw. Learning Syst.}, pp. 1–-21, 2020.

\bibitem{Sorzano2014}
C. O. S. Sorzano, J. Vargas, and A. P. Montano,
``A survey of dimensionality reduction techniques,''
\emph{arXiv:1403.2877 [cs, q-bio, stat]}, Mar. 2014, Accessed: Apr. 03, 2020. [Online]. Available: http://arxiv.org/abs/1403.2877.

\bibitem{Bengio2014}
Y. Bengio, A. Courville, and P. Vincent,
``Representation Learning: A Review and New Perspectives,''
\emph{arXiv:1206.5538 [cs]}, Apr. 2014, Accessed: Apr. 04, 2020. [Online]. Available: http://arxiv.org/abs/1206.5538.


\bibitem{Rossi2017}
R. A. Rossi, R. Zhou, and N. K. Ahmed,
``Deep Feature Learning for Graphs,''
\emph{arXiv:1704.08829 [cs, stat]}, Oct. 2017, Accessed: Apr. 04, 2020. [Online]. Available: http://arxiv.org/abs/1704.08829.

\bibitem{Cao2016}
S. Cao, W. Lu, and Q. Xu,
``Deep neural networks for learning graph representations,''
\emph{Proceedings of the Thirtieth AAAI Conference on Artificial Intelligence}, pp. 1145–-1152, 2016.

\bibitem{Škrlj2019}
B. Škrlj, J. Kralj, J. Konc, M. Robnik-Šikonja, and N. Lavrač,
``Deep Node Ranking: Structural Network Embedding and End-to-End Node Classification,''
\emph{arXiv:1902.03964 [cs, stat]}, Apr. 2019, Accessed: Apr. 04, 2020. [Online]. Available: http://arxiv.org/abs/1902.03964.

\bibitem{Min2010}
M. Min, L. van der Maaten, Z. Yuan, A. Bonner, Z. Zhang,
``Deep Supervised t-Distributed Embedding,''
\emph{Proceedings of the 27th International Conference on Machine Learning (ICML-10)}, pp. 791--798, 2010.

\bibitem{Pan2019}
S. Pan, R. Hu, G. Long, J. Jiang, L. Yao, and C. Zhang,
``Adversarially Regularized Graph Autoencoder for Graph Embedding,''
\emph{arXiv:1802.04407 [cs, stat]}, Jan. 2019, Accessed: Feb. 29, 2020. [Online]. Available: http://arxiv.org/abs/1802.04407.

\bibitem{Maaten2008}
L.J.P. van der Maaten and G.E. Hinton.
"Visualizing High-Dimensional Data Using t-SNE", 
{\em Journal of Machine Learning Research}, pp. 2579-2605, 2008.

\bibitem{Maaten2014}
L.J.P. van der Maaten.
"Accelerating t-SNE using Tree-Based Algorithms", 
{\em Journal of Machine Learning Research}, pp. 3221-3245, 2014.

\bibitem{Leow2019}
Y. Y. Leow, T. Laurent, and X. Bresson,
``GraphTSNE: A Visualization Technique for Graph-Structured Data,''
\emph{arXiv:1904.06915 [cs, stat]}, Apr. 2019, Accessed: Feb. 29, 2020. [Online]. Available: http://arxiv.org/abs/1904.06915.

\bibitem{Mao2015}
Q. Mao, L. Wang, S. Goodison, and Y. Sun,
``Dimensionality Reduction Via Graph Structure Learning,''
\emph{Proceedings of the 21th ACM SIGKDD International Conference on Knowledge Discovery and Data Mining - KDD `15}, pp. 765-–774, 2015.

\bibitem{Goyal2018-a}
P. Goyal, N. Kamra, X. He, and Y. Liu,
``DynGEM: Deep Embedding Method for Dynamic Graphs,''
\emph{arXiv:1805.11273 [cs]}, May 2018, Accessed: Apr. 04, 2020. [Online]. Available: http://arxiv.org/abs/1805.11273.

\bibitem{Yan2007}
S. Yan, D. Xu, B. Zhang, H. Zhang, Q. Yang, and S. Lin,
``Graph Embedding and Extensions: A General Framework for Dimensionality Reduction,''
\emph{IEEE Trans. Pattern Anal. Mach. Intell.}, vol. 29, no. 1, pp. 40–-51, 2007.

\bibitem{Goyal2018-b}
P. Goyal and E. Ferrara,
``Graph embedding techniques, applications, and performance: A survey,''
\emph{Knowledge-Based Systems}, vol. 151, pp. 78–-94, 2018.

\bibitem{Kruiger2017}
J. F. Kruiger, P. E. Rauber, R. M. Martins, A. Kerren, S. Kobourov, and A. C. Telea,
``Graph Layouts by t‐SNE,''
\emph{Knowledge-Based Systems}, vol. 36, no. 3, pp. 283–294, 2017.

\bibitem{Simonovsky2018}
M. Simonovsky and N. Komodakis,
``GraphVAE: Towards Generation of Small Graphs Using Variational Autoencoders,''
\emph{Artificial Neural Networks and Machine Learning – ICANN 2018, vol. 11139, V. Kůrková, Y. Manolopoulos, B. Hammer, L. Iliadis, and I. Maglogiannis, Eds. Cham: Springer International Publishing}, pp. 412–-422, 2018.

\bibitem{Epasto2019}
A. Epasto and B. Perozzi,
``Is a Single Embedding Enough? Learning Node Representations that Capture Multiple Social Contexts,''
\emph{The World Wide Web Conference - WWW ’19}, pp. 394–-404, 2019.

\bibitem{Yu2018}
W. Yu et al.,
``Learning Deep Network Representations with Adversarially Regularized Autoencoders,''
\emph{Proceedings of the 24th ACM SIGKDD International Conference on Knowledge Discovery \& Data Mining}, pp. 2663–-2671, 2018.

\bibitem{Grover2016}
A. Grover and J. Leskovec,
``node2vec: Scalable Feature Learning for Networks,''
\emph{Proceedings of the 22nd ACM SIGKDD International Conference on Knowledge Discovery and Data Mining - KDD `16}, pp. 855–-864, 2016.

\bibitem{Wang2016}
D. Wang, P. Cui, and W. Zhu,
``Structural Deep Network Embedding,''
\emph{Proceedings of the 22nd ACM SIGKDD International Conference on Knowledge Discovery and Data Mining - KDD `16}, pp. 1225–-1234, 2016.

\bibitem{Perozzi2014}
B. Perozzi, R. Al-Rfou, and S. Skiena,
``DeepWalk: Online Learning of Social Representations,''
\emph{Proceedings of the 20th ACM SIGKDD international conference on Knowledge discovery and data mining - KDD `14}, pp. 701–-710, 2014.

\bibitem{Cao2015}
S. Cao, W. Lu, and Q. Xu,
``Grarep: Learning graph representations with global structural information,''
\emph{Proceedings of the 24th ACM International on Conference on Information and Knowledge Management}, pp. 891–-900, 2015.

\bibitem{Tang2015}
J. Tang, M. Qu, M. Wang, M. Zhang, J. Yan, and Q. Mei,
``LINE: Large-scale Information Network Embedding,''
\emph{Proceedings of the 24th International Conference on World Wide Web - WWW `15}, pp. 1067–-1077, 2015.

\bibitem{Hamilton2018}
W. L. Hamilton, R. Ying, and J. Leskovec,
``Representation Learning on Graphs: Methods and Applications,''
\emph{arXiv:1709.05584 [cs]}, Apr. 2018, Accessed: Apr. 04, 2020. [Online]. Available: http://arxiv.org/abs/1709.05584.

\bibitem{Yang2016}
Z. Yang, W. W. Cohen, and R. Salakhutdinov,
``Revisiting Semi-Supervised Learning with Graph Embeddings,''
\emph{Proceedings of The 33rd International Conference on Machine Learning}, vol. 48, pp. 40--48, 2016.

\bibitem{Gan2018}
J. Gan, X. Zhu, C. Lei, Y. Li, H. Yu, and S. Zhang,
``Robust Graph Dimensionality Reduction,''
\emph{Proceedings of the Twenty-Seventh International Joint Conference on Artificial Intelligence}, pp. 3257–-3263, 2018.

\bibitem{Ghosh2020}
T. Ghosh and M. Kirby,
``Supervised Dimensionality Reduction and Visualization using Centroid-encoder,''
\emph{arXiv:2002.11934 [cs, stat]}, Feb. 2020, Accessed: Apr. 04, 2020. [Online]. Available: http://arxiv.org/abs/2002.11934.

\bibitem{Gutiérrez-Gómez2019}
L. Gutiérrez-Gómez and J.-C. Delvenne,
``Unsupervised Network Embedding for Graph Visualization, Clustering and Classification,''
\emph{arXiv:1903.05980 [[cs, stat]}, Mar. 2019, Accessed: Apr. 03, 2020. [Online]. Available: http://arxiv.org/abs/1903.05980.

\bibitem{Kipf2016}
T. N. Kipf and M. Welling,
``Variational Graph Auto-Encoders,''
\emph{NIPS Workshop on Bayesian Deep Learning}, 2016.

\bibitem{Abu-El-Haija2018}
S. Abu-El-Haija, B. Perozzi, R. Al-Rfou, and A. A. Alemi,
``Watch Your Step: Learning Node Embeddings via Graph Attention,''
\emph{Proceedings of the 32nd International Conference on Neural Information Processing Systems}, pp. 9198–-9208, 2018.


\bibitem{Yuan2020}
Y. Yuan,
``Introduction to Deep Learning for Graphs and Where It May Be Heading,''
\emph{https://syncedreview.com}, Feb. 20 2020, Accessed: Apr. 04, 2020. [Online]. Available: https://syncedreview.com/2020/02/20/introduction-to-deep-learning-for-graphs-and-where-it-may-be-heading/.

\end{thebibliography}

\end{document}